\documentclass{article}

 \usepackage[preprint]{neurips_2026}

\usepackage{natbib}

\usepackage[utf8]{inputenc} 
\usepackage[T1]{fontenc}    
\usepackage{url}            
\usepackage{booktabs}       
\usepackage{amsfonts}       
\usepackage{nicefrac}       
\usepackage{microtype}      
\usepackage{xcolor}         
\usepackage{multirow}
\usepackage{adjustbox}
\usepackage{subcaption}
\usepackage{amsmath}

\usepackage{graphicx}
\usepackage{capt-of} 

\usepackage[pagebackref=false, breaklinks=true, colorlinks, citecolor=citecolor, linkcolor=linkcolor, bookmarks=false]{hyperref}
\definecolor{citecolor}{HTML}{0071BC}
\definecolor{linkcolor}{HTML}{ED1C24}

\usepackage{colortbl}
\colorlet{best}{blue!15}
\colorlet{second}{blue!5}

\renewcommand{\medskip}{\vspace{0.01in}}
\renewcommand{\paragraph}[1]{{\noindent\textbf{#1}}}

\newcommand{\methodname}{{SearchCast}}
\title{How Good Can Linear Models Be \\
for Time-Series Forecasting?}

\author{%
  Lang Huang$^{1, 2}$ \quad
  Jinglue Xu$^{1}$ \quad
  Luke Darlow$^{1}$ \\
  $^{1}$Sakana AI, Tokyo, Japan \\
  $^{2}$National Institute of Informatics, Japan\\
  \{langhuang,~jingluexu,~luke\}@sakana.ai \\
}

\begin{document}

\maketitle

\vspace{-0.2in}
\begin{abstract}
Time-series forecasting research has been moving steadily toward larger architectures, from specialized transformers to general-purpose foundation models, on the assumption that capacity is what unlocks accuracy. We take the opposite position: most of the gap can be closed at far lower cost by tuning preprocessing rather than scaling models.  We use Ridge regression~\cite{hoerl1970ridge} as the testbed, since it has a closed-form solution and interpretable weights, which let the optimal hyperparameters be read off the search directly. We search over context length, local normalization, regularization, and augmentation on eight standard benchmarks and find three patterns. (1) Optimal lookback is strongly series-specific and often non-monotonic in forecast horizon, with fitted power-law exponents ranging from $+0.46$ on ETTm2 to $-0.19$ on Exchange and Traffic, challenging the convention that longer horizons need longer history. (2) Normalizing over a learned trailing fraction of the context, rather than its entirety, is almost universally preferred. (3) Series within the same dataset often disagree on hyperparameters; the optimal degree of cross-series sharing varies from fully shared to fully per-series. The resulting models beat prior linear forecasters on most dataset-horizon entries and exceed Transformer, MLP, and CNN baselines on six of eight benchmarks. The optimized hyperparameters also serve as a diagnostic on the data itself, revealing structures that larger models absorb silently into their learned parameters. We provide an accompanying \href{https://sakanaai.github.io/SearchCast/}{interactive online demonstration} and the \href{https://github.com/SakanaAI/SearchCast}{codebase}.
\end{abstract}



\begin{figure}[h]
  \centering
  \includegraphics[width=\textwidth]{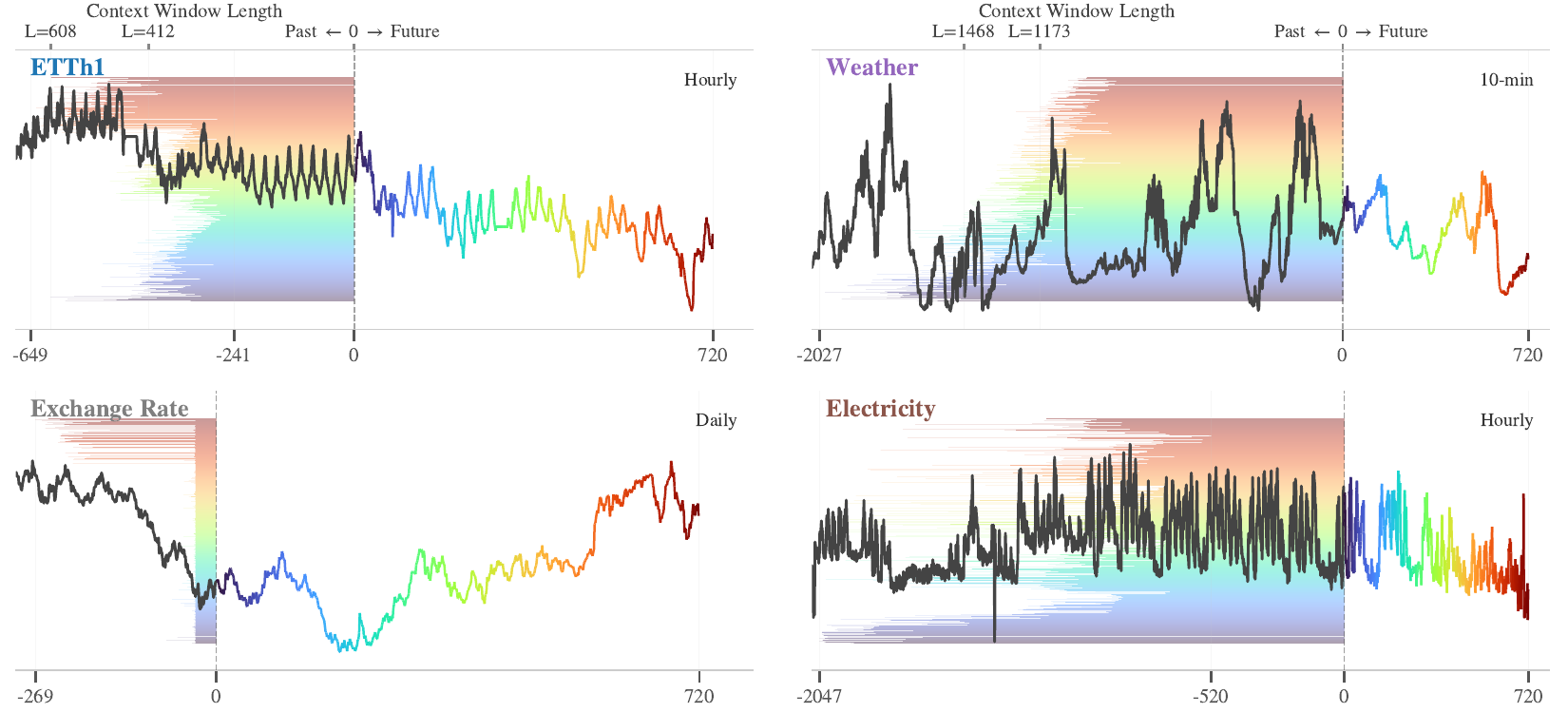}
  \vspace{-0.2in}
  \caption{Per-horizon optimal context-horizon relationships for four time series. The context lengths were obtained via hyperparameter search and are shown as color-matched bars.
  }
  \label{fig:teaser_series_w_cont_wind}
\end{figure}

\vspace{-0.3in}
\section{Introduction}
\label{sec:intro}
\vspace{-0.2in}
The long-term time series forecasting literature has followed a familiar arc over the past several years. Transformers~\cite{vaswani2017attention} were adapted to the task with progressively more sophisticated attention and patching mechanisms~\citep{wu2021autoformer,zhou2022fedformer,nie2022patchtst,liu2023itransformer}, Multi-Layer Perceptron (MLP) and Convolutional Neural Network (CNN) alternatives demonstrated that comparable accuracy could be achieved with simpler inductive biases~\citep{wu2022timesnet,wang2024timemixer,chen2023tsmixer,das2023long}, and linear models entered the conversation when \cite{zeng2023dlinear} showed that a single linear layer, applied after trend-seasonal decomposition, could outperform several modern transformers. The resulting proliferation of linear variants, \emph{e.g.}, DLinear, NLinear~\cite{zeng2023dlinear}, RLinear\cite{li2023rlinear}, SparseTSF~\cite{lin2024sparsetsf}, gave the impression of a rich design space, until \cite{toner2024analysis} proved that these models are functionally equivalent to unconstrained linear regression over suitably augmented features, and that closed-form ordinary least squares solutions can match or exceed their SGD-trained counterparts. This collapse of seemingly architectural diversity into a single model class raises a natural question: if the model is effectively fixed, where should the remaining degrees of freedom be spent?

We argue that the answer is preprocessing. The standard evaluation protocol in time series forecasting fixes context length, normalization strategy, and data augmentation to a single setting per benchmark and then compares architectures. This convention makes sense when the goal is to isolate the effect of model design, but it systematically disadvantages models whose capacity is too limited to absorb suboptimal input representations through learned parameters. Transformers with millions of parameters can partially compensate for a poorly chosen lookback window or an uninformative normalization scheme; a linear model cannot. The result is that linear methods appear weaker than they are, not because they lack expressive power for the task, but because they are more sensitive to choices that are rarely tuned.
We test this hypothesis with Ridge regression~\cite{hoerl1970ridge}, applied to a systematic search over context length, local normalization windows, regularization strength, and augmentation in both time and frequency domains across eight standard benchmarks at per-horizon and per-series granularity. Ridge has a closed-form solution, no hidden nonlinearity, runs a trial in a few milliseconds on a GPU, and produces weights that can be inspected directly. The same transparency that makes it a competitive forecaster also makes it a diagnostic instrument: the structure of the optimal hyperparameters reveals properties of the data that deeper models would absorb silently into their learned representations.

\begin{figure}[b]
  \vspace{-0.3in}
  \centering
  \includegraphics[width=.42\textwidth]{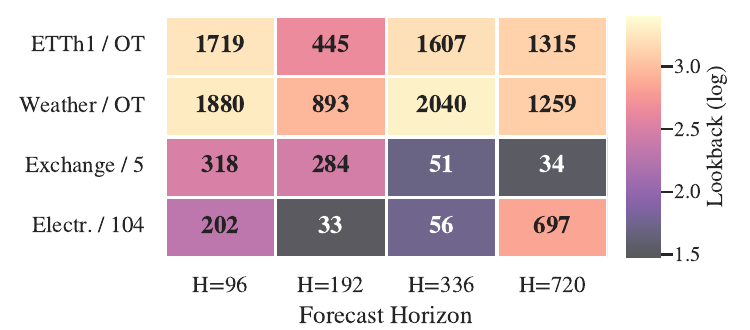}
  \hfill
  \includegraphics[width=.56\textwidth]{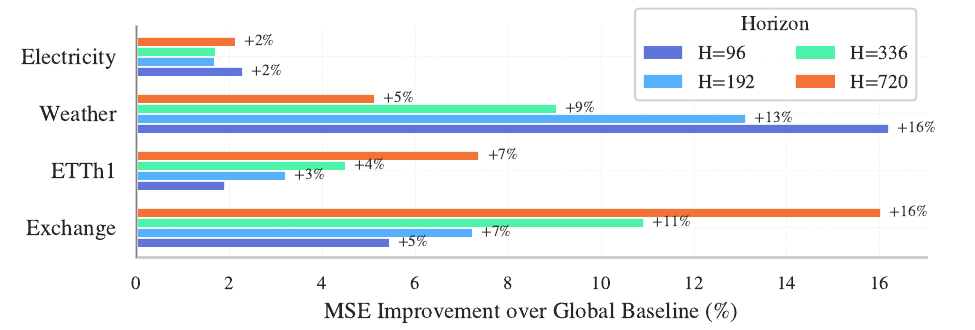}
  \vspace{-0.05in}
  \caption{\textit{Left}: Median optimal lookback per (series, horizon) varies. \textit{Right}: Adapting context per horizon yields up to +16\% MSE improvement over a global baseline across these four datasets.}
  \label{fig:teaser_bc}
  \vspace{-0.2in}
\end{figure}

The search reveals three intriguing observations. First, the relationship between optimal lookback and forecast horizon is strongly dataset-specific (Figures~\ref{fig:teaser_series_w_cont_wind}\&\ref{fig:teaser_bc}) and often non-monotonic (Figure~\ref{fig:lookback_scaling}). When fitting a power-law $L^* = a\cdot H^b$ for the searched optimal lookback $L^*$ and the prediction horizon $H$, the exponent $b$ ranges from $+0.46$ on ETTm2 to $-0.19$ on Exchange and Traffic, contradicting the common assumption that longer horizons demand longer history~\cite{oreshkin2020nbeats,challu2023nhits}. Second, normalizing over a learned trailing fraction of the context, rather than its entirety as in prior work, consistently improves accuracy, indicating that recent local statistics carry more signal than global ones (Figure~\ref{fig:hp_heatmaps}). Third, series within the same dataset often prefer different hyperparameters (Figure~\ref{fig:series_group_size}), and the optimal degree of cross-series sharing varies from fully shared on ETTh1 to fully per-series on Weather, suggesting the heterogeneity across channels as an underexplored axis in forecasting.

Our contributions are as follows. First, we show that carefully tuned Ridge regression outperforms prior linear forecasters across most datasets and time horizons, and matches or exceeds Transformer- and MLP-based architectures on six of eight benchmarks, while being orders of magnitude cheaper to train. Second, we demonstrate that the optimal hyperparameter landscape of a transparent linear model encodes structural properties of time series data (scaling behavior, normalization preferences, series heterogeneity) that are informative beyond the model itself and can guide the design of more complex forecasters. Third, we release \methodname{}~(\href{https://github.com/SakanaAI/SearchCast}{URL}), a reproducible pipeline that supports per-horizon and per-series search with configurable ablation controls, facilitating future studies to diagnose their own datasets with the same methodology.
\section{Related Work}
\label{sec:related_work}
\paragraph{Transformer-based forecasters.} 
Attention-based architectures dominated long-term forecasting for several years. Autoformer~\citep{wu2021autoformer} and FEDformer~\citep{zhou2022fedformer} replaced standard self-attention with frequency-domain operators to capture seasonal structure. PatchTST~\citep{nie2022patchtst} segmented inputs into patches and processed each channel independently, and remains the strongest transformer baseline on long-term benchmarks. iTransformer~\citep{liu2023itransformer} inverted the attention axis to model cross-channel dependencies. Recent benchmarks~\citep{qiu2024tfb,toner2024analysis} find that no single transformer wins across datasets and horizons, which raises the question of where the remaining gains should come from.

\paragraph{MLP and CNN alternatives.} A parallel line of work explored simpler architectures. TimesNet~\citep{wu2022timesnet} reshapes 1D series into 2D tensors and applies convolutions. TiDE~\citep{das2023long} and TimeMixer~\citep{wang2024timemixer} are encoder-decoder MLPs, the latter mixing across temporal resolutions. TSMixer~\citep{chen2023tsmixer} reaches competitive accuracy with an attention-free MLP mixer.

\paragraph{Linear models.} DLinear~\citep{zeng2023dlinear} showed that a trend-seasonal decomposition followed by two linear layers could outperform mordern transformers; NLinear, RLinear~\citep{li2023rlinear}, and SparseTSF~\citep{lin2024sparsetsf} extended the recipe with last-value normalization, reversible instance normalization~\citep{kim2022revin}, and parameter-sparse forecasting, respectively. Toner and Darlow~\cite{toner2024analysis} then proved that DLinear, NLinear, and several related variants are functionally equivalent to unconstrained linear regression over augmented features, and that closed-form ordinary least squares (OLS) solutions match or beat the same models trained with SGD. This collapses the diverse linear forecasters into a single model class.

\paragraph{Foundation models for time series.} A more recent line proposes general-purpose forecasters trained on large corpora and applied zero-shot, including Chronos~\citep{ansari2024chronos}, TimesFM~\citep{abhimanyu2024decoder}, Moirai~\citep{woo2024unified}, and DAM~\cite{darlow2025dam}. These models target broad coverage rather than per-dataset accuracy, and on standard supervised benchmarks.
However, reaching this generality requires large pretraining corpora and substantially more compute than per-dataset fitting, while accuracy on standard supervised benchmarks is often comparable to tuned per-dataset baselines, a small marginal return on the added cost.

\paragraph{Positioning.} 
We build on the unification of~\citep{toner2024analysis}. Instead of proposing another linear variant, we fix the model to Ridge regression~\cite{hoerl1970ridge} and spend the remaining budget on preprocessing: context length, local normalization, regularization, and augmentation, searched per-horizon and per-series across eight benchmarks. This closes most of the gap to deeper baselines and turns the optimized hyperparameters into a diagnostic: how optimal context scales with horizon, whether a trailing fraction of the window beats the full window for normalization, and how much variates within a dataset disagree, all readable off the search and mostly invisible in the parameters of deeper models.
\section{Method}
\label{sec:method}

\subsection{Preliminary: Ridge Regression}

Toner and Darlow~\cite{toner2024analysis} showed that recent linear forecasting variants are functionally equivalent to unconstrained linear regression over appropriately augmented feature sets. We take this unification as our starting point and build on the simple and efficient setup: Ridge regression~\cite{hoerl1970ridge} with a closed-form solution. Given a context window $\mathbf{x} \in \mathbb{R}^{L}$ of $L$ past observations for a single variate and a forecast horizon $H$, the model predicts $\hat{\mathbf{y}} = \mathbf{W}\mathbf{x} + \mathbf{b}$ where $\mathbf{W} \in \mathbb{R}^{H \times L}$ and $\mathbf{b} \in \mathbb{R}^{H}$. The weight matrix is obtained by solving
\begin{equation}
    \mathbf{W}^* = \arg\min_{\mathbf{W}} \|\mathbf{Y} - \mathbf{W}\mathbf{X}\|_F^2 + \alpha \|\mathbf{W}\|_F^2
\end{equation}
where $\mathbf{X} \in \mathbb{R}^{L\times N}$ and $\mathbf{Y}\in \mathbb{R}^{H\times N}$ are the training context and target matrices assembled from $N$ sliding windows, respectively; and $\alpha$ is the regularization strength. The solution $\mathbf{W}^* = \mathbf{Y}\mathbf{X}^\top(\mathbf{X}\mathbf{X}^\top + \alpha \mathbf{I})^{-1}$ is computed in closed form without iterative optimization, making training effectively instantaneous on a single GPU and highly amenable to large-scale hyperparameter search. Each variate is modeled independently following the channel-independent paradigm that has proven effective across both linear~\citep{zeng2023dlinear} and Transformer-based~\citep{nie2022patchtst} forecasters. 

\subsection{Preprocessing Pipeline}

The preprocessing applied before the simple Ridge regression is where most of the modeling flexibility resides. We parameterize it along four axes.

\paragraph{Normalization.} 
We consider two normalization scopes. Global normalization computes statistics over the entire training set for each variate; local normalization computes them over the most recent $r \cdot L$ time steps of each input window, where $r \in (0, 1]$ is the \emph{local ratio}. In either case we support two methods: standardization (subtract mean, divide by standard deviation) and robust normalization (subtract median, divide by interquartile range $[0.25, 0.75]$). Local normalization with $r = 1$ recovers full-window instance normalization as used in prior work~\citep{toner2024analysis}; values $r < 1$ restrict the statistics to a trailing fraction of the context, allowing the model to adapt to the most recent distributional regime. The local ratio is searched on a log scale between $0.001$ and $1.0$. Following~\cite{toner2024analysis}, under local normalization we append the per-window scale $\sigma$ as an additional feature in place of the intercept (which is uninformative once the context has zero local mean), letting the regression learn a volatility-dependent shift; a zero coefficient on $\sigma$ recovers pure instance normalization.

\paragraph{Context length.} 
The lookback $L$ determines how much history the model sees and directly controls the dimensionality of the regression problem. We search $L \in [32, 2048]$ on a log scale, with upper bounds capped by the data size and forecast horizon.

\paragraph{Augmentation.} We optionally perturb training windows with additive noise scaled by $\sigma$ in either the time (Gaussian noise) or the frequency domain (Gaussian perturbation of Fourier coefficients), where $\sigma$ is searched on a log scale between $0.001$ and $0.5$. A third option applies no augmentation at all.

\paragraph{Regularization.} Rather than selecting a single $\alpha$, we evaluate a grid of 21 logarithmically spaced values (in range $[10^{-6}, 10^{3}]$) and retain the one that minimizes validation loss, effectively treating regularization as an inner optimization loop that imposes no additional search cost.

\subsection{Grouped Hyperparameter Search}
Prior work almost always fixes a single set of preprocessing hyperparameters across every benchmark, horizon step, and variate. This is computationally convenient but assumes a degree of homogeneity within the datasets that rarely holds in practice: different horizons/variates can demand different context lengths. The opposite extreme, an independent search per horizon step and per variate, would capture this heterogeneity but is rarely attempted because the search budget grows combinatorially. We introduce a grouped search scheme that moves continuously between these two endpoints.

\paragraph{Horizon grouping.} We partition the $H$ forecast steps into contiguous blocks of size $g_h$ and share hyperparameters within each block. Setting $g_h = 1$ recovers fully per-step tuning; setting $g_h = H$ recovers the global baseline. Intermediate values allow hyperparameters to vary smoothly across the forecast horizon without requiring a separate search per step. 

\paragraph{Series grouping.} Similarly, the $C$ variates are partitioned into groups of size $g_s$. Setting $g_s = 1$ yields a fully per-series search that can capture heterogeneous dynamics across channels; setting $g_s = C$ shares \emph{weights and hyperparameters} across all variates. 

\paragraph{Unified view.} The pair $(g_h, g_s)$ defines a grid of hyperparameter search cells over the horizon$\times$series space. Each cell runs an independent Optuna search with a Tree-structured Parzen Estimator~\citep{akiba2019optuna} over the joint preprocessing space: lookback $L$, normalization scope and method, local ratio $r$, augmentation type and intensity $\sigma$, with $\alpha$ selected by inner grid search for each trial. The global model ($g_h = H, g_s = C$) and fully local model ($g_h = 1, g_s = 1$) are special cases of this framework.

\paragraph{Cross-validation.} Each Optuna trial evaluates candidate hyperparameters using $k$-fold expanding-window cross-validation on the training set, with folds constructed chronologically to respect temporal ordering. We use $k = 3$ in all main experiments. The best hyperparameters are then applied to the held-out test set, which is never used during the search.
\begin{table*}[t]
\centering
\small
\caption{Long-term multivariate forecasting results (MSE). Left panel: linear models; right panel: nonlinear models. OLS, FITS, and DLinear report the results with instance normalization from \cite{toner2024analysis}. \colorbox{best}{\textbf{Bold}}: best per row. \colorbox{second}{\underline{Underline}}: second best.}
\label{tab:main_results}
\resizebox{\textwidth}{!}{%
\begin{tabular}{cc|cccc|ccccc}
\toprule
 & & \multicolumn{4}{c|}{\textbf{Linear Models}} & \multicolumn{5}{c}{\textbf{Nonlinear Models}} \\
\cmidrule(lr){3-6} \cmidrule(lr){7-11}
Dataset & $H$ & \textbf{\methodname{}} & OLS & FITS & DLinear & PatchTST & iTransformer & TimeMixer & TimesNet & Autoformer \\
\midrule
\multirow{5}{*}{\rotatebox{90}{ETTm1}}
& 96 & \cellcolor{best}\textbf{0.297} & \cellcolor{second}\underline{0.307} & 0.309 & 0.312 & 0.329 & 0.334 & 0.320 & 0.338 & 0.505 \\
& 192 & \cellcolor{best}\textbf{0.332} & \cellcolor{second}\underline{0.336} & 0.338 & 0.341 & 0.367 & 0.377 & 0.361 & 0.374 & 0.553 \\
& 336 & \cellcolor{best}\textbf{0.357} & \cellcolor{second}\underline{0.365} & 0.367 & 0.372 & 0.401 & 0.426 & 0.408 & 0.410 & 0.621 \\
& 720 & \cellcolor{best}\textbf{0.396} & \cellcolor{second}\underline{0.415} & 0.417 & 0.422 & 0.456 & 0.491 & 0.469 & 0.478 & 0.671 \\
\cmidrule(lr){2-11}
& \textit{Avg} & \cellcolor{best}\textit{\textbf{0.346}} & \cellcolor{second}\textit{\underline{0.356}} & \textit{0.358} & \textit{0.362} & \textit{0.388} & \textit{0.407} & \textit{0.389} & \textit{0.400} & \textit{0.588} \\
\midrule
\multirow{5}{*}{\rotatebox{90}{ETTm2}}
& 96 & \cellcolor{best}\textbf{0.160} & \cellcolor{second}\underline{0.162} & \cellcolor{second}\underline{0.162} & 0.163 & 0.175 & 0.180 & 0.175 & 0.187 & 0.255 \\
& 192 & \cellcolor{best}\textbf{0.212} & \cellcolor{second}\underline{0.216} & 0.217 & 0.217 & 0.241 & 0.250 & 0.241 & 0.249 & 0.281 \\
& 336 & \cellcolor{best}\textbf{0.256} & \cellcolor{second}\underline{0.268} & 0.269 & 0.269 & 0.305 & 0.311 & 0.305 & 0.321 & 0.339 \\
& 720 & \cellcolor{best}\textbf{0.323} & \cellcolor{second}\underline{0.349} & 0.350 & 0.354 & 0.402 & 0.412 & 0.378 & 0.408 & 0.433 \\
\cmidrule(lr){2-11}
& \textit{Avg} & \cellcolor{best}\textit{\textbf{0.237}} & \cellcolor{second}\textit{\underline{0.249}} & \textit{0.250} & \textit{0.251} & \textit{0.281} & \textit{0.288} & \textit{0.275} & \textit{0.291} & \textit{0.327} \\
\midrule
\multirow{5}{*}{\rotatebox{90}{ETTh1}}
& 96 & \cellcolor{best}\textbf{0.369} & \cellcolor{second}\underline{0.375} & 0.377 & 0.379 & 0.414 & 0.386 & \cellcolor{second}\underline{0.375} & 0.384 & 0.449 \\
& 192 & \cellcolor{best}\textbf{0.400} & 0.413 & 0.413 & 0.419 & 0.460 & 0.441 & \cellcolor{second}\underline{0.405} & 0.436 & 0.500 \\
& 336 & \cellcolor{best}\textbf{0.423} & 0.445 & \cellcolor{second}\underline{0.432} & 0.451 & 0.501 & 0.487 & 0.439 & 0.491 & 0.521 \\
& 720 & \cellcolor{second}\underline{0.430} & 0.460 & \cellcolor{best}\textbf{0.428} & 0.470 & 0.507 & 0.503 & 0.469 & 0.521 & 0.514 \\
\cmidrule(lr){2-11}
& \textit{Avg} & \cellcolor{best}\textit{\textbf{0.405}} & \textit{0.423} & \cellcolor{second}\textit{\underline{0.412}} & \textit{0.430} & \textit{0.471} & \textit{0.454} & \textit{0.422} & \textit{0.458} & \textit{0.496} \\
\midrule
\multirow{5}{*}{\rotatebox{90}{ETTh2}}
& 96 & \cellcolor{best}\textbf{0.268} & \cellcolor{second}\underline{0.270} & \cellcolor{second}\underline{0.270} & 0.275 & 0.302 & 0.297 & 0.289 & 0.340 & 0.346 \\
& 192 & \cellcolor{best}\textbf{0.330} & \cellcolor{second}\underline{0.331} & \cellcolor{second}\underline{0.331} & 0.342 & 0.388 & 0.380 & 0.333 & 0.402 & 0.456 \\
& 336 & \cellcolor{best}\textbf{0.354} & \cellcolor{best}\textbf{0.354} & \cellcolor{best}\textbf{0.354} & 0.359 & 0.426 & 0.428 & 0.374 & 0.452 & 0.482 \\
& 720 & 0.383 & \cellcolor{second}\underline{0.380} & \cellcolor{best}\textbf{0.377} & 0.384 & 0.431 & 0.427 & 0.416 & 0.462 & 0.515 \\
\cmidrule(lr){2-11}
& \textit{Avg} & \cellcolor{second}\textit{\underline{0.334}} & \cellcolor{second}\textit{\underline{0.334}} & \cellcolor{best}\textit{\textbf{0.333}} & \textit{0.340} & \textit{0.387} & \textit{0.383} & \textit{0.353} & \textit{0.414} & \textit{0.450} \\
\midrule
\multirow{5}{*}{\rotatebox{90}{Electricity}}
& 96 & \cellcolor{second}\underline{0.130} & 0.133 & 0.133 & 0.134 & \cellcolor{best}\textbf{0.129} & 0.148 & 0.150 & 0.168 & 0.201 \\
& 192 & \cellcolor{best}\textbf{0.145} & 0.148 & 0.148 & 0.149 & \cellcolor{second}\underline{0.147} & 0.162 & 0.163 & 0.184 & 0.222 \\
& 336 & \cellcolor{best}\textbf{0.161} & 0.164 & 0.164 & 0.165 & \cellcolor{second}\underline{0.163} & 0.178 & 0.178 & 0.198 & 0.231 \\
& 720 & \cellcolor{second}\underline{0.199} & 0.203 & 0.203 & 0.205 & \cellcolor{best}\textbf{0.197} & 0.225 & 0.220 & 0.220 & 0.254 \\
\cmidrule(lr){2-11}
& \textit{Avg} & \cellcolor{best}\textit{\textbf{0.159}} & \textit{0.162} & \textit{0.162} & \textit{0.163} & \cellcolor{best}\textit{\textbf{0.159}} & \textit{0.178} & \textit{0.178} & \textit{0.192} & \textit{0.227} \\
\midrule
\multirow{5}{*}{\rotatebox{90}{Traffic}}
& 96 & \cellcolor{second}\underline{0.379} & 0.385 & 0.386 & 0.387 & \cellcolor{best}\textbf{0.360} & 0.395 & 0.392 & 0.593 & 0.613 \\
& 192 & \cellcolor{second}\underline{0.391} & 0.397 & 0.398 & 0.399 & \cellcolor{best}\textbf{0.379} & 0.417 & 0.408 & 0.617 & 0.616 \\
& 336 & \cellcolor{second}\underline{0.404} & 0.410 & 0.411 & 0.412 & \cellcolor{best}\textbf{0.392} & 0.433 & 0.422 & 0.629 & 0.622 \\
& 720 & \cellcolor{second}\underline{0.441} & 0.448 & 0.449 & 0.450 & \cellcolor{best}\textbf{0.432} & 0.467 & 0.459 & 0.640 & 0.660 \\
\cmidrule(lr){2-11}
& \textit{Avg} & \cellcolor{second}\textit{\underline{0.404}} & \textit{0.410} & \textit{0.411} & \textit{0.412} & \cellcolor{best}\textit{\textbf{0.391}} & \textit{0.428} & \textit{0.420} & \textit{0.620} & \textit{0.628} \\
\midrule
\multirow{5}{*}{\rotatebox{90}{Weather}}
& 96 & \cellcolor{best}\textbf{0.141} & \cellcolor{best}\textbf{0.141} & \cellcolor{second}\underline{0.142} & \cellcolor{second}\underline{0.142} & 0.149 & 0.174 & 0.164 & 0.172 & 0.266 \\
& 192 & \cellcolor{best}\textbf{0.184} & \cellcolor{best}\textbf{0.184} & \cellcolor{second}\underline{0.185} & \cellcolor{second}\underline{0.185} & 0.194 & 0.221 & 0.215 & 0.219 & 0.307 \\
& 336 & \cellcolor{best}\textbf{0.234} & \cellcolor{best}\textbf{0.234} & 0.236 & \cellcolor{second}\underline{0.235} & 0.245 & 0.278 & 0.272 & 0.280 & 0.359 \\
& 720 & \cellcolor{best}\textbf{0.304} & \cellcolor{second}\underline{0.307} & \cellcolor{second}\underline{0.307} & 0.310 & 0.314 & 0.358 & 0.351 & 0.365 & 0.419 \\
\cmidrule(lr){2-11}
& \textit{Avg} & \cellcolor{best}\textit{\textbf{0.215}} & \cellcolor{second}\textit{\underline{0.217}} & \textit{0.218} & \textit{0.218} & \textit{0.225} & \textit{0.258} & \textit{0.250} & \textit{0.259} & \textit{0.338} \\
\midrule
\multirow{5}{*}{\rotatebox{90}{Exchange}}
& 96 & \cellcolor{best}\textbf{0.081} & 0.086 & 0.087 & \cellcolor{second}\underline{0.085} & 0.088 & 0.086 & 0.086 & 0.107 & 0.197 \\
& 192 & \cellcolor{best}\textbf{0.167} & 0.180 & 0.183 & 0.178 & \cellcolor{second}\underline{0.176} & 0.177 & 0.177 & 0.226 & 0.300 \\
& 336 & \cellcolor{second}\underline{0.305} & 0.343 & 0.344 & 0.335 & \cellcolor{best}\textbf{0.301} & 0.331 & 0.338 & 0.367 & 0.509 \\
& 720 & \cellcolor{best}\textbf{0.811} & 0.992 & 0.965 & 0.920 & \cellcolor{second}\underline{0.901} & \cellcolor{second}\underline{0.901} & 0.920 & 0.964 & 1.447 \\
\cmidrule(lr){2-11}
& \textit{Avg} & \cellcolor{best}\textit{\textbf{0.341}} & \textit{0.400} & \textit{0.395} & \textit{0.380} & \cellcolor{second}\textit{\underline{0.366}} & \textit{0.374} & \textit{0.380} & \textit{0.416} & \textit{0.613} \\
\bottomrule
\end{tabular}%
}
\vspace{-0.1in}
\end{table*}

\section{Experiments}
\label{sec:exp}

\subsection{Experimental Setup}
\paragraph{Datasets}
We evaluate on eight widely used multivariate forecasting benchmarks spanning diverse domains and temporal granularities: ETTh1 and ETTh2 (electricity transformer temperature, hourly), ETTm1 and ETTm2 (same source, 15-minute intervals), Weather (21 meteorological indicators, 10-minute), Electricity (321 clients, hourly), Traffic (862 road sensors, hourly), and Exchange (daily exchange rates of 8 countries). Following standard protocol~\citep{wu2021autoformer,zeng2023dlinear,toner2024analysis}, each dataset is split chronologically into training, validation, and test sets at a 6:2:2 ratio for ETT and 7:1:2 for the others. 

\paragraph{Evaluation Protocol.} We report MSE at horizons $H \in \{96, 192, 336, 720\}$ and per-dataset averages. Search uses 20 Optuna trials per cell, and 3-fold cross-validation. Because our method searches the context length $L$, each validation fold includes the additional pre-validation history needed to form the input window, but keeps the number of validation target points fixed across all candidate $L$. Thus longer contexts do not receive more validation samples nor a different validation interval.

\paragraph{Baselines.} \textbf{Linear:} OLS, FITS (with instance normalization, from~\cite{toner2024analysis}), and DLinear~\cite{zeng2023dlinear}. These isolate the gain attributable to preprocessing search rather than model capacity. \textbf{Nonlinear:} PatchTST~\cite{nie2022patchtst}, iTransformer~\citep{liu2023itransformer}, TimeMixer~\citep{wang2024timemixer}, TimesNet~\citep{wu2022timesnet}, and Autoformer~\citep{wu2021autoformer}, spanning Transformer, MLP, and CNN families. Nonlinear baselines use the best lookback $L$ as published; linear baselines from~\cite{toner2024analysis} use $L=720$. Our method searches over $L$ (while keeping the validation samples fixed for fair comparisons), a degree of freedom we argue is underexplored.

\subsection{Main Results}
Table~\ref{tab:main_results} reports MSE across the eight benchmarks. \textbf{Within the linear class}, our pipeline \methodname{} achieves the best average MSE on seven of eight datasets and ties FITS on ETTh2 ($0.334$ vs.\ $0.333$). Reductions over OLS reach $4.8\%$ on ETTm2, $4.3\%$ on ETTh1, and $14.8\%$ on Exchange. The gap is consistent across $H$ rather than driven by a single horizon, indicating the gain comes from preprocessing choices that generalize. \textbf{Against nonlinear baselines}, ours wins the average on six of eight benchmarks despite having no nonlinearity and no learned representation. Both losses are to PatchTST: Electricity (tied at $0.159$) and Traffic ($0.404$ vs.\ $0.391$), the two largest datasets ($321$ and $862$ channels), where shared representations across similar series plausibly help Transformers more than simple Ridge. This is consistent with Figure~\ref{fig:series_group_size}, where moderate series grouping is favorable for these two datasets. On the remaining six datasets, the margin over PatchTST ranges from $4\%$ (Weather) to $16\%$ (ETTm2). The linear comparison isolates the contribution of preprocessing search; the nonlinear comparison shows it is large enough to close the gap on most benchmarks.


\begin{figure}[t]
    \centering
    \vspace{-0.1in}
    \includegraphics[width=\linewidth]{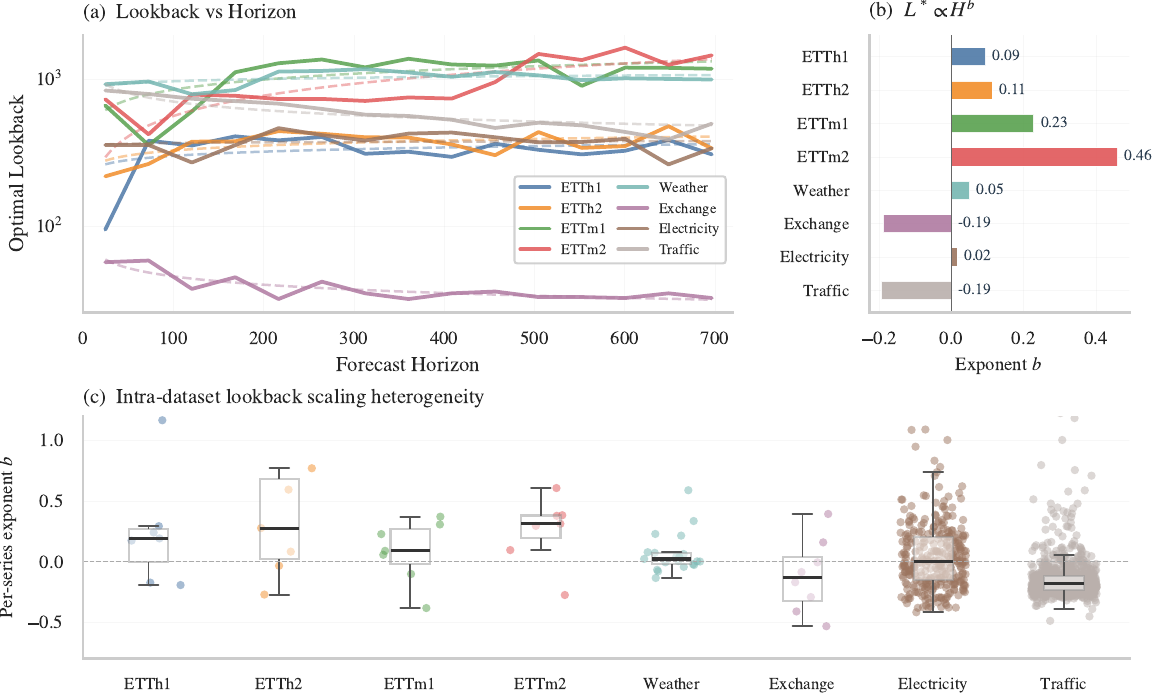}
    \caption{Optimal lookback $L^*$ vs.\ forecast horizon $H$ across 8 datasets. (a) Median lookback (log-log) with IQR bands and power-law fits $L^* \propto H^b$. (b) Fitted exponent $b$ per dataset. (c) Per-series exponent distribution within each dataset.}
    \label{fig:lookback_scaling}
    \vspace{-0.1in}
\end{figure}

\section{Analysis}
\label{sec:analysis}
We use the search output as a diagnostic of the data. We first study how optimal lookback changes with horizon and series (\S\ref{sec:lookback_scaling}), then how much hyperparameter sharing is safe across series and horizons (\S\ref{sec:series_grouping}). Forecasts and refit Ridge weights connect these choices to model behavior (\S\ref{sec:qualitative}). The remaining subsections isolate the main preprocessing choices: local normalization (\S\ref{sec:local_norm}), series-level variation in $r$ and $\alpha$ (\S\ref{sec:series_heterogeneity}), and augmentation type/intensity (\S\ref{sec:augmentation}).

\subsection{Lookback \emph{v.s.} Horizon Across Datasets}
\label{sec:lookback_scaling}

Figure~\ref{fig:lookback_scaling}(a,b) fits $L^* = a \cdot H^b$ to the searched optimal context at each horizon group \textbf{per dataset}. Exponents span both signs and an order of magnitude: from $b = +0.46$ on ETTm2 down to $b = -0.19$ on Exchange and Traffic. ETTm2 is the only dataset where the conventional intuition holds strongly, there longer horizons ask for more history, though sublinearly. ETTm1 and ETTh2 sit in a mild positive regime ($b \approx 0.1$--$0.2$); Weather and Electricity are nearly flat ($|b| < 0.1$), settling on $\sim 10^3$ steps regardless of horizon. The negative exponents on Exchange and Traffic are the most informative: longer horizons prefer \emph{shorter} context, consistent with non-stationarity where distant history actively misleads the model. The standard $L = 96$ default is therefore wrong in two opposite directions at once: it underserves Weather and Electricity by roughly an order of magnitude, and overserves Exchange and Traffic at long horizons, where the optimum drops below $96$.

Figure~\ref{fig:lookback_scaling}(c) repeats the fit \textbf{per variate}. Dataset-level exponents are aggregates over considerable intra-dataset spread. Weather is the only benchmark whose per-series exponents cluster tightly around the median; ETTh2, Electricity, and Traffic straddle zero, meaning some channels prefer growing context with horizon while others prefer shrinking it. This highlights that a single shared lookback is insufficient even within one dataset, which motivates the series/horizon grouping analysis below.

\begin{figure}[t]
    \centering
    \includegraphics[width=\linewidth]{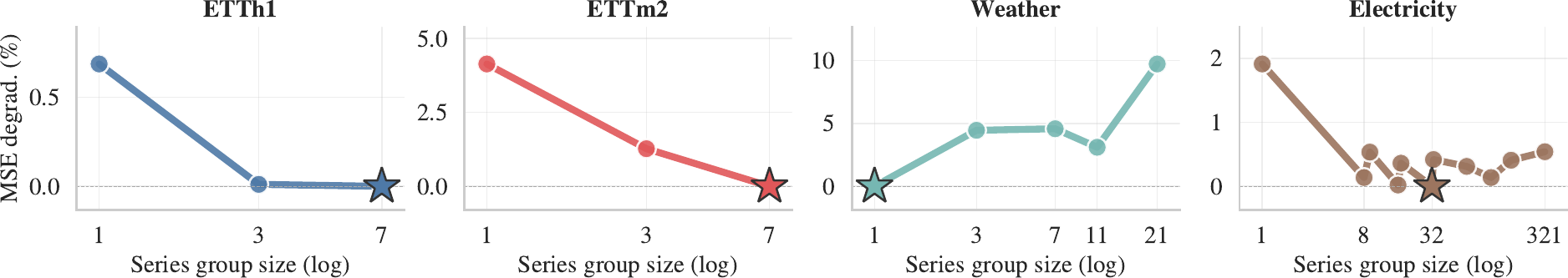}
    \vskip 0.1in
    \includegraphics[width=\linewidth]{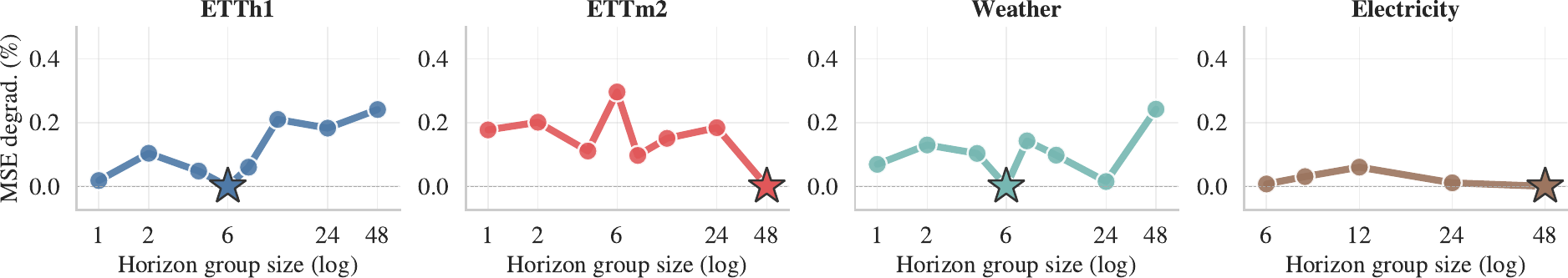}
    \caption{Effect of series (\emph{Top}) and horizon (\emph{Bottom}) grouping on forecasting accuracy across 4 datasets. Each panel shows MSE degradation (\%) relative to the best group size (marked with $\star$) as a function of group size, from per-series/horizon to fully shared.}
    \label{fig:series_group_size}
\end{figure}

\subsection{Series and Horizon Grouping}
\label{sec:series_grouping}

Figure~\ref{fig:series_group_size} (top) sweeps the series group size $g_s$ from per-series ($g_s = 1$) to fully shared ($g_s = S$), reporting MSE degradation relative to the per-dataset best. The picture is non-monotone. \textbf{Fully shared} is best on both ETT benchmarks, where per-series search degrades MSE by $0.5$--$4\%$, suggesting physically related variates regularize each other. \textbf{Fully per-series} is best on Weather, where degradation grows monotonically with $g_s$ to $10\%$ at $g_s = 21$, suggesting its 21 channels (pressure, humidity, wind, rainfall) live on incompatible scales. Electricity prefers \textbf{intermediate} sizes ($g_s = 32$). The result indicates that the optimal degree of cross-series sharing is a property of the dataset.

The bottom row sweeps $g_h \in \{1, 2, 3, 4, 6, 8, 12, 16, 24, 48\}$. $g_h$ groups \emph{consecutive} horizons into bins of width $g_h$ that share one HP setting: at $g_h = 1$, horizons $H = 95$ and $H = 96$ are tuned separately; at $g_h = 48$, all of $H = 49,\ldots,96$ share a single setting. The four evaluation cutoffs $\{96, 192, 336, 720\}$ are far enough apart that they land in different bins for every valid $g_h$. So as $g_h$ grows, the four reported cutoffs keep their own tuned HPs; only the horizons \emph{near} each cutoff are progressively absorbed into its bin. The curves are flat ($\leq 0.4\%$ degradation), with optima at $g_h = 48$ on ETTm2 and Electricity, and $g_h = 6$ on ETTh1 and Weather. This says HPs vary smoothly with $H$ within a $\sim$48-step neighborhood, consistent with the smooth $L^*(H)$ trends in §\ref{sec:lookback_scaling}. We adopt $g_h = 48$ by default for its efficiency and spend the trial budget on searching $g_s$ instead.

\subsection{Forecasts and Learned Weights Visualization}
\label{sec:qualitative}

\begin{figure}[t]
    \centering
    \includegraphics[width=\linewidth]{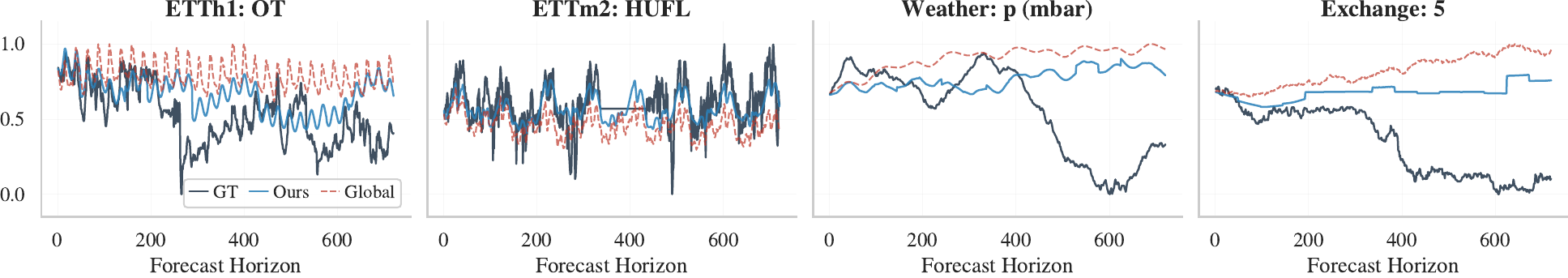}
    \vspace{-0.15in}
    \caption{Forecast comparison. Our method (blue) closely tracks the ground truth (black), while the global baseline (red) reverts to the mean, particularly on non-stationary series (Weather, Exchange). The choppiness in Weather and especially Exchange is an artifact of switching hyperparameters across forecast settings, which can create discontinuities between segments.}
    \label{fig:forecast_comparison}
\end{figure}

Figure~\ref{fig:forecast_comparison} shows the visible effect of the search. With fixed dataset-level defaults, Ridge drifts toward the mean as the horizon grows, especially on Weather and Exchange. The tuned model keeps following the slow drift because these datasets select short effective context, small local-normalization windows, or both (Figures~\ref{fig:lookback_scaling}b and~\ref{fig:hp_heatmaps}).

\begin{figure}[t]
    \centering
    \includegraphics[width=\linewidth]{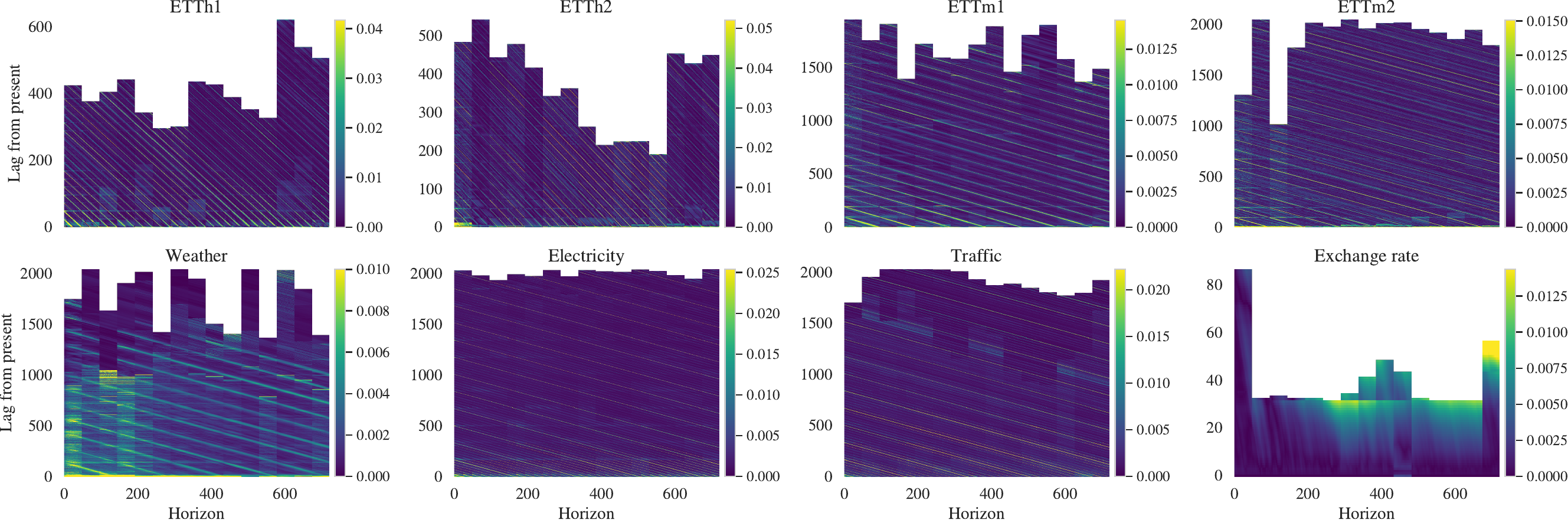}
    \vspace{-0.15in}
    \caption{Weight magnitude $|w|$ over lag and forecast horizon. Lighter shades indicate larger magnitudes. Lag is measured from the most recent input (bottom row); white regions lie beyond each model's chosen lookback.}
    \label{fig:lag_horizon_weights}
\end{figure}

Figure~\ref{fig:lag_horizon_weights} shows what the selected lookbacks are used for. Exchange puts almost all weight on recent lags, matching its short-memory behavior. Traffic and ETTh2 use calendar lags: Traffic has strong weekly bands near $168, 336, \ldots$, while ETTh2 shows a shorter daily comb. Electricity mixes lag-$0/1$ dependence with remote anchors. Weather and ETTm datasets often place their largest weights hundreds of steps back. Thus a long $L^*$ mainly gives the model access to a few phase-matched observations, not to the whole history. The jumps at $g_h {=}48$ bin boundaries show the tradeoff: horizon grouping barely changes MSE ($\leq 0.4\%$; \S\ref{sec:series_grouping}), but it can change the weights abruptly.

\subsection{Global \emph{v.s.} Local Normalization}
\label{sec:local_norm}
\begin{figure*}[t]
    \vspace{-0.1in}
    \centering
    \includegraphics[width=\textwidth]{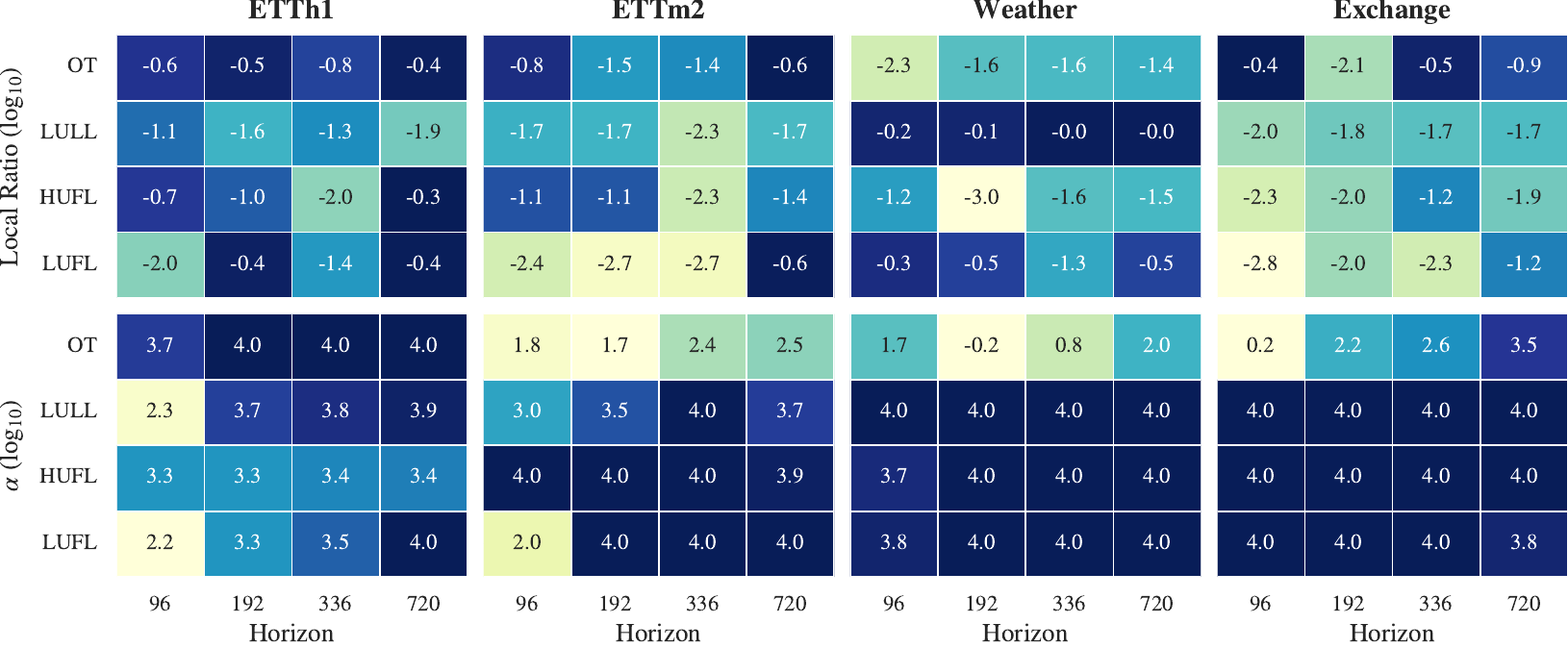}
    \vspace{-0.2in}
    \caption{Per-series hyperparameter (local ratio $r$ and regularization $\alpha$) heatmaps, with $g_h = 48$.}
    \label{fig:hp_heatmaps}
    \vspace{-0.1in}
\end{figure*}

Prior linear forecasters normalize either globally (training-set statistics) or locally (full input window)~\citep{toner2024analysis,zeng2023dlinear}. We relax the local case to a learned trailing fraction $r \in (0,1]$: $r = 1$ recovers full-window normalization, $r < 1$ restricts statistics to the last $r \cdot L$ steps. Across ETTh1, ETTm2, Weather, and Exchange, we found that local is selected in 62--100\% of dataset--horizon cells; the local-ratio column of Figure~\ref{fig:hp_heatmaps} shows that the optimal $\log_{10} r$ is almost always strictly negative, clustered in $[-2.5, -0.5]$ — trailing fractions between $0.3\%$ and $30\%$ of the window. \textbf{Full-window normalization is essentially never selected.} The effect is strongest on ETTm2 and Exchange, where many cells sit near $\log_{10} r = -2$: only the final few percent of the lookback set the scale. These series are nonstationary on the scale of $L$, and full-window statistics blur regimes the model is better off keeping distinct. We also tested a robust median/IQR variant of local normalization, but it underperformed mean/std in all tested settings, suggesting that tail variation and large deviations carry useful scale information in these benchmarks. We therefore use local standardization in Table~\ref{tab:main_results} and search only the trailing fraction $r$.

\subsection{Hyperparameter Variations Across Series}
\label{sec:series_heterogeneity}

Figures~\ref{fig:lookback_scaling} and~\ref{fig:hp_heatmaps} identify \emph{which} hyperparameters drive the Weather \emph{vs.}\ ETTh1 contrast. Lookback is uniform across series on ETTh1 and dispersed on Weather (Figure~\ref{fig:lookback_scaling}c). The local-ratio and $\alpha$ heatmaps in Figure~\ref{fig:hp_heatmaps} sharpen the picture. \textbf{ETTh1:} both rows are visibly uniform across the four variates and vary more with horizon than with series, consistent with variates from one physical system. \textbf{Weather:} the two rows diverge — local ratio varies widely (OT at $\log_{10} r \approx -2$ vs.\ LULL near $0$), and $\alpha$ shows a large gap between OT ($\log_{10}\alpha \approx 1\text{--}2$) and the other channels (near $4$). The per-series gain on Weather (Figure~\ref{fig:series_group_size}) therefore comes from local ratio and $\alpha$, not lookback. ETTm2 and Exchange sit between these extremes; grouping by measurement type or a few clusters would likely capture most of the heterogeneity at a fraction of full per-series cost.

\begin{figure*}[t]
    \centering
    \includegraphics[width=\textwidth]{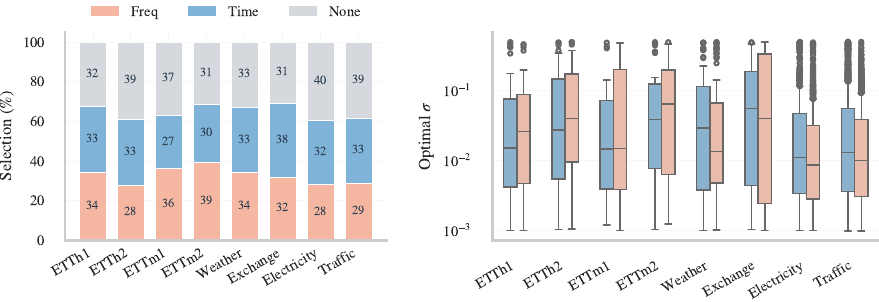}
    \vspace{-0.2in}
    \caption{Augmentation analysis. \emph{Left}: Proportion of trials selecting frequency-domain noise, time-domain noise, or no augmentation. \emph{Right}: Distribution of optimal $\sigma$ conditional on augmentation being selected, broken down by domain.}
    \label{fig:augmentation_analysis}
    \vspace{-0.2in}
\end{figure*}

\subsection{Augmentation Selection and Intensity}
\label{sec:augmentation}

Figure~\ref{fig:augmentation_analysis}(a): augmentation is selected in $60$--$70\%$ of horizon groups on every benchmark, with time- and frequency-domain noise split roughly evenly (time slightly more common). Figure~\ref{fig:augmentation_analysis}(b) displays \emph{how} they are selected.
The within-dataset spread of optimal $\sigma$ is itself dataset-dependent: Weather and Exchange are widest (near- and far-horizon groups demand different noise levels); ETTm1, Electricity, and Traffic are tighter.

\section{Conclusion}
\label{sec:conclusion}

The reputation of linear models as uncompetitive baselines in long-term forecasting reflects under-tuned preprocessing, not limited model capacity. A Ridge regression searched over context length, local normalization, regularization, and augmentation matches or exceeds prior linear baselines as well as Transformer, MLP, and CNN baselines on most standard benchmarks, at a fraction of the training cost. Beyond accuracy, the optimized hyperparameters serve as a diagnostic lens on the data: optimal lookback can grow, plateau, or shrink with the forecast horizon depending on dataset stationarity; normalization is almost always local, restricted to a trailing fraction of the window; cross-series sharing is dataset-specific, ranging from fully shared to fully per-series, and the heterogeneity is driven by normalization and regularization rather than lookback; and the same locality and seasonality the search recovers in the hyperparameters are visible directly in the trained weights. These preprocessing choices transfer to any model class, and the released \methodname{} pipeline makes the same diagnostic cheap to run on new datasets.

{
\small
\bibliographystyle{abbrv}
\bibliography{main.bib}
}

\newpage
\appendix

\section{Long-range linear autocorrelation in the benchmark series}
\label{sec:app_acf}

The per-dataset search of Section~\ref{sec:lookback_scaling} selects context lengths that differ by two orders of magnitude, from a few dozen steps on Exchange to roughly a thousand on Weather and Electricity (Figure~\ref{fig:lookback_scaling}). This appendix examines a basic question behind that result. When the search assigns the model a long context, is there genuinely useful information that far in the past, or is the model only re-representing an obvious daily or weekly cycle that a much simpler seasonal method~\citep{zeng2023dlinear,zhou2022fedformer} could capture without a long context? We address it by measuring how strongly each series is linearly related to its own past, both on the raw series and after \emph{deseasonalization}. Deseasonalization here means removing a series' known periodic cycles (for instance the daily and weekly ones), so that any remaining correlation cannot be attributed to those cycles alone. We apply four deseasonalization methods of increasing strength (defined below) and check whether the long-range correlation survives all of them.

The \emph{autocorrelation} at lag $k$ is the Pearson correlation between the series at time $t$ and the same series $k$ steps earlier, $\rho(y_t, y_{t-k})$, computed over all $t$ and then averaged across the channels of a dataset~\citep{hyndman2021forecasting}. It ranges from $-1$ to $1$. A value near $0$ means the value $k$ steps ago carries essentially no \emph{linear} information about the present, while a large magnitude means it carries substantial such information. Because we focus on the linear relationship, autocorrelation is precisely the kind of structure our forecaster can exploit. Before measuring, each series is z-scored using training-set statistics only, that is, shifted and scaled to mean $0$ and variance $1$ using quantities computed on the training portion alone, so that the measurement is comparable across series and does not use test data.

From weakest to strongest, we apply four deseasonalization methods. The \emph{raw} variant removes nothing. The \emph{per-position-in-period} variant subtracts the average value at each position of the dominant cycle (for example the mean for each hour-of-day). The \emph{harmonic} variant subtracts fitted sine and cosine waves at all known periods such as daily and weekly. The \emph{STL} variant applies a standard procedure that splits a series into a trend, a seasonal part, and a remainder, keeping only the remainder~\citep{cleveland1990stl}. The rationale for trying all four is that if a long-lag correlation were entirely a periodic cycle, a model that explicitly represents cycles, such as the trend-seasonal decomposition inside DLinear~\citep{zeng2023dlinear} or the frequency-domain operators of Autoformer and FEDformer~\citep{wu2021autoformer,zhou2022fedformer}, could reproduce it at low cost, and a long context would offer a purely linear model no genuine advantage. Each reported correlation comes with a $95\%$ confidence interval from a pair-resampling bootstrap. We resample the observed $(y_t, y_{t-k})$ pairs with replacement ($B = 200$ times), recompute the correlation each time, and report the middle $95\%$ of those values. A narrow interval means the estimate is statistically reliable.

\begin{figure}[h]
\centering
\includegraphics[width=\textwidth]{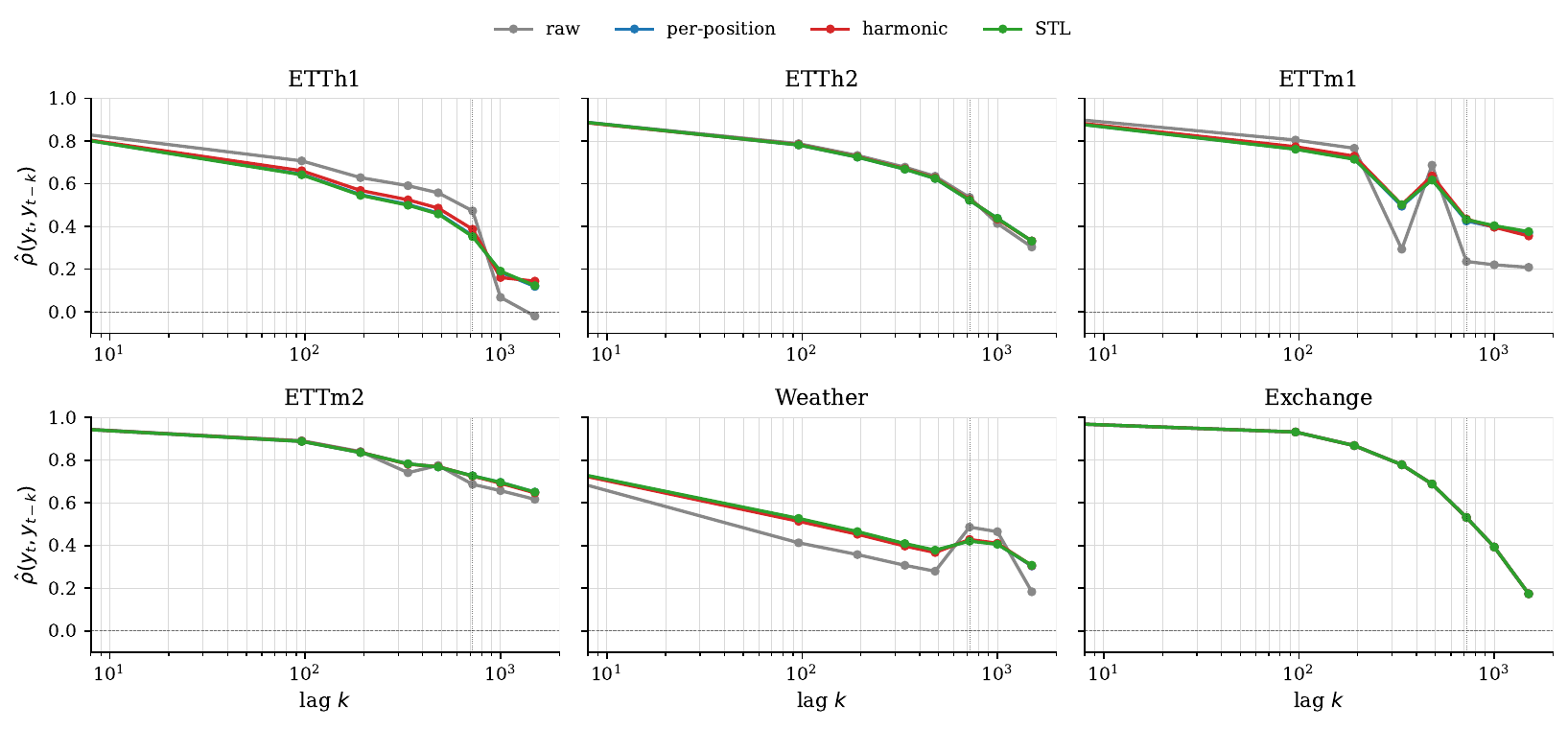}
\caption{Channel-averaged Pearson autocorrelation $\hat\rho(y_t, y_{t-k})$ as a function of lag $k$, for the six standard benchmarks under four deseasonalization schemes. Dotted vertical line marks $k = 720$. The confidence intervals are narrow (typical $95\%$ bootstrap half-width below $0.01$, given in Table~\ref{tab:acf720_app} for $k = 720$). The four schemes agree at long lags, with all six datasets retaining $\hat\rho \ge 0.39$ at $k = 720$ after harmonic deseasonalization.}
\label{fig:app_acf_decay}
\end{figure}

Figure~\ref{fig:app_acf_decay} traces the autocorrelation across all measured lags for the four variants, and Table~\ref{tab:acf720_app} reports the slice at lag $k = 720$, the longest forecast horizon evaluated in Section~\ref{sec:exp}. All six datasets retain a substantial correlation at $k = 720$, between $0.39$ and $0.73$, with confidence intervals far above $0$. The four removal methods agree closely at long lags, so the surviving correlation is not an artifact of any single deseasonalization choice. On ETTm1 and ETTm2 the deseasonalized correlation is in fact \emph{higher} than the raw one, because removing the strong daily cycle uncovers slower structure that the cycle had been masking.

\begin{table}[h]
\centering
\small
\caption{Long-range linear autocorrelation in the six standard benchmarks, computed at lag $k = 720$ after harmonic deseasonalization at all configured periods. Values are channel-averaged Pearson correlations on the z-scored concatenation of the train, validation and test splits. Confidence intervals are 95\% pair-resampling bootstraps with $B = 200$.}
\label{tab:acf720_app}
\begin{tabular}{lcc}
\toprule
Dataset & $\hat\rho(y_t, y_{t-720})$ & 95\% CI \\
\midrule
ETTh1          & 0.387 & [0.381, 0.394] \\
ETTh2          & 0.526 & [0.521, 0.531] \\
ETTm1          & 0.435 & [0.433, 0.438] \\
ETTm2          & 0.725 & [0.723, 0.728] \\
Weather        & 0.428 & [0.400, 0.451] \\
Exchange       & 0.532 & [0.527, 0.538] \\
\bottomrule
\end{tabular}
\end{table}

Even after the obvious daily and weekly cycles are removed, a value from $720$ steps in the past still carries a real and moderately strong linear association with the current value (correlation $0.39$ to $0.73$). Therefore, when the search provides the linear model with a long context on these datasets, the model is genuinely exploiting long-range information rather than re-representing a cycle that a simpler seasonal method could capture.

\section{How much accuracy comes from context length alone}
\label{sec:app_lookback}

Section~\ref{sec:lookback_scaling} shows \emph{which} context length the full search prefers on each dataset. This appendix isolates a narrower question. Of the accuracy our method gains, how much is due to the context length $L$ \emph{by itself}, as opposed to the other preprocessing choices (normalization, regularization, augmentation)? To answer it we hold every other preprocessing setting fixed and vary only $L$.

We use a single fixed preprocessing configuration everywhere, which we call the universal default. This default uses local mean normalization, the standard regularization grid described in Section~\ref{sec:method}, and no augmentation. Keeping this configuration identical across datasets means that any change in error can be attributed to $L$ alone. We sweep $L \in \{24, 48, 96, 192, 336, 480, 720, 1000, 1500, 2000\}$, forecast at the longest horizon $H = 720$, and report the test mean squared error (MSE, the held-out squared prediction error used throughout our evaluation in Section~\ref{sec:exp}, where lower is better). For each $L$ we report the median MSE across channels and random seeds, the median being robust to a few unusually hard channels.

\begin{figure}[h]
\centering
\includegraphics[width=0.9\textwidth]{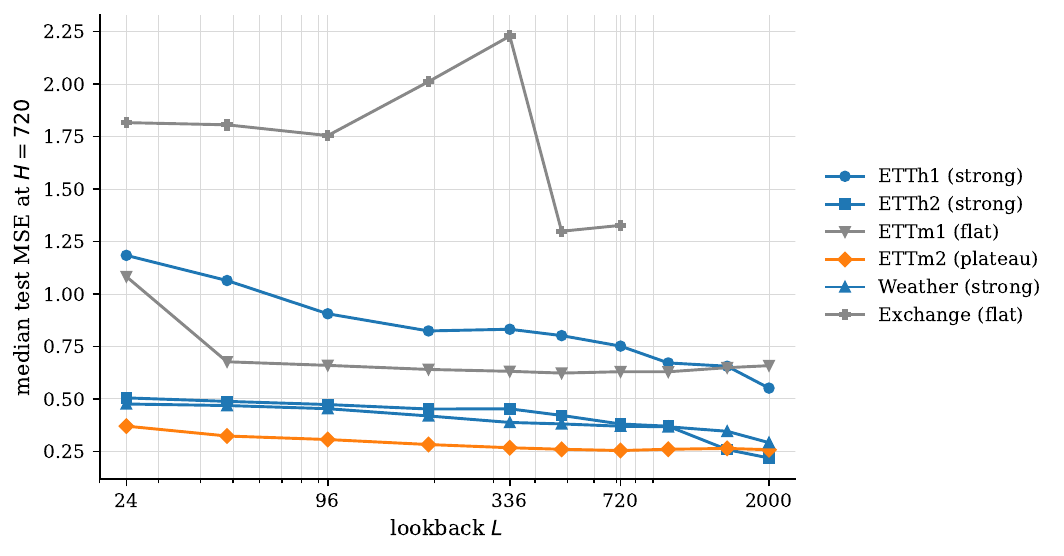}
\caption{Median per-channel test MSE on Ridge with universal-default preprocessing at $H = 720$ as a function of the lookback $L$. Lines are color-coded by regime. Strong (blue) continues to improve up to $L = 2000$, plateau (orange) saturates around $L = 192$--$336$, and flat (gray) is essentially insensitive to $L$.}
\label{fig:app_lookback_curves}
\end{figure}

Figure~\ref{fig:app_lookback_curves} plots the full error-versus-$L$ curves and Table~\ref{tab:lookback_h720} lists the endpoints. The datasets fall into three clear regimes. In the \emph{strong} regime (ETTh1, ETTh2, Weather) the error continues to decrease as $L$ grows to $2000$, for a total reduction between $36\%$ and $53\%$. In the \emph{plateau} regime (ETTm2) most of the reduction is achieved by $L \approx 192$--$336$, after which the curve saturates, for a $16\%$ reduction. In the \emph{flat} regime (ETTm1) additional context yields almost no improvement. Exchange is shown only partially because its test split is too short to form windows with both $L = 2000$ and $H = 720$. Two points follow. First, on the strong-regime datasets a fixed short default such as $L = 96$ forfeits most of the attainable accuracy before any other preprocessing setting is tuned. Second, which regime a dataset falls into cannot be predicted from its metadata in advance and must instead be measured, which is why our method (Section~\ref{sec:method}) searches $L$ per dataset rather than fixing it. We also note that re-running the full preprocessing search while holding $L$ fixed at a short value does not recover the long-context gain, which indicates that context length itself, rather than the other settings, is the source of the improvement.

\begin{table}[h]
\centering
\small
\caption{Median per-channel test MSE on Ridge with universal-default preprocessing at $H = 720$ on the six standard benchmarks, as a function of the lookback $L$. The total-drop column is the percent change between $L = 96$ and $L = 2000$, with negative values indicating improvement. Exchange is flagged because its test split has insufficient samples to form $L = 2000$, $H = 720$ windows.}
\label{tab:lookback_h720}
\begin{tabular}{lcccl}
\toprule
Dataset & $L = 96$ & $L = 720$ & $L = 2000$ & Total drop \\
\midrule
ETTh1          & 0.91 & 0.75 & 0.55 & $-39\%$ \\
ETTh2          & 0.47 & 0.38 & 0.22 & $-53\%$ \\
Weather        & 0.45 & 0.37 & 0.29 & $-36\%$ \\
ETTm2          & 0.31 & 0.25 & 0.26 & $-16\%$ \\
ETTm1          & 0.66 & 0.63 & 0.66 & flat \\
Exchange       & 1.75 & 1.33 & --   & partial \\
\bottomrule
\end{tabular}
\end{table}

To confirm that the model performs substantive forecasting rather than merely repeating recent values, we compare it against a \emph{persistence} forecast, the naive rule that predicts the future to equal the last observed value, $\hat{y}_{t+h} = y_t$ (also called a no-change or random-walk forecast)~\citep{hyndman2021forecasting}. Table~\ref{tab:ar1_h720} reports the ratio of the persistence error to the Ridge error at $H = 720$. Ridge is between $1.4$ and $5.9$ times more accurate on every dataset.

\begin{table}[h]
\centering
\small
\caption{Persistence (last-value) baseline at $H = 720$. The persistence forecast predicts $\hat{y}_{t+h} = y_t$. Ridge MSE is the universal-default Ridge from Table~\ref{tab:lookback_h720} at the per-dataset best $L \in \{24, \ldots, 2000\}$. The ratio is how many times smaller the Ridge error is.}
\label{tab:ar1_h720}
\begin{tabular}{lccc}
\toprule
Dataset & Persistence MSE & Ridge MSE & Ratio (Persistence / Ridge) \\
\midrule
ETTh1          & 1.10 & 0.43 & 2.56 \\
ETTh2          & 0.60 & 0.38 & 1.56 \\
ETTm1          & 2.33 & 0.40 & 5.87 \\
ETTm2          & 0.46 & 0.32 & 1.44 \\
Weather        & 0.55 & 0.30 & 1.81 \\
Exchange       & 1.42 & 0.81 & 1.75 \\
\bottomrule
\end{tabular}
\end{table}

For some datasets (ETTh1, ETTh2, Weather), extending the linear model's context, with no other change, reduces its error by $36$ to $53\%$. For others (ETTm1) it yields almost no improvement. No single context length is best for every dataset, so the only reliable way to identify each dataset's optimal value is to search for it (Section~\ref{sec:method}). The persistence comparison confirms that the model performs genuine forecasting rather than reproducing the most recent value.

\section{Does a nonlinear model capture structure the linear model misses?}
\label{sec:app_bds}

Our main results (Section~\ref{sec:exp}, Table~\ref{tab:main_results}) show that the tuned linear model matches or beats the strongest nonlinear baseline, PatchTST, on six of the eight benchmarks. A natural concern is that MSE may fail to reflect everything a model captures. Perhaps a nonlinear model captures genuine structure that does not register in the error value, so that the linear model would only appear competitive. This appendix addresses that concern directly by examining the prediction errors that each model produces.

The \emph{residuals} of a model are its prediction errors at the forecast step, $y_{\text{true}} - \hat{y}$. If a model has captured everything predictable in a series, its residuals should resemble random noise with no remaining pattern. A linear model can only represent linear relationships (weighted sums of past values), so any \emph{nonlinear} structure (a pattern in which the influence of the past is not a fixed linear weight) remains in its residuals. We therefore assess how closely each model's residuals resemble pure noise, using the Brock--Dechert--Scheinkman (BDS) test~\citep{brock1996bds}, a standard statistical test of whether a sequence is independent random noise or still contains residual dependence, as implemented in \texttt{statsmodels}~\citep{seabold2010statsmodels}. We apply it at embedding dimension two, meaning it inspects how often pairs of points fall close together relative to the frequency expected under pure randomness. A larger BDS statistic indicates more residual non-random structure. We compare the statistic itself rather than its $p$-value, because with this many points every $p$-value is driven to $0$ and can no longer distinguish the models.

For each model we form the per-cell difference $\Delta = \mathrm{BDS}_{\text{arch}} - \mathrm{BDS}_{\text{Ridge}}$, matched on the same (dataset, context length, horizon, seed, channel). A \emph{negative} $\Delta$ means that model's residuals are closer to noise than the linear model's, that is, it captured nonlinear structure that the linear model did not. As in Appendix~\ref{sec:app_acf}, each comparison carries a $95\%$ confidence interval from resampling, and we call a cell \emph{significant} when that interval excludes $0$ (for example, $[-3.0, -1.2]$ does, whereas $[-1.5, +0.8]$ does not). Table~\ref{tab:bds_app} aggregates the comparison over the six standard benchmarks (ETTh1, ETTh2, ETTm1, ETTm2, Weather, Exchange) and, separately, over the two large many-channel datasets (Electricity with $321$ channels and Traffic with $862$), where Section~\ref{sec:exp} reports that PatchTST tends to achieve lower error (Table~\ref{tab:main_results}). DLinear, itself a linear model, lies near $\Delta \approx 0$, as expected, and serves as a control confirming that the test does not report differences where none exist.

\begin{table}[h]
\centering
\small
\caption{Paired BDS comparison against Ridge. $\Delta = \mathrm{BDS}_{\mathrm{arch}} - \mathrm{BDS}_{\mathrm{Ridge}}$ is the per-cell difference of the BDS statistic at embedding dimension $2$, paired by $(\text{dataset}, L, H, \text{seed}, \text{channel})$ and bootstrapped with $B = 5{,}000$ resamples. A cell is significant when its paired 95\% confidence interval excludes zero, and a negative mean $\Delta$ means the architecture leaves less non-random structure than Ridge. The \emph{standard} group is the six benchmarks ETTh1, ETTh2, ETTm1, ETTm2, Weather and Exchange. The \emph{large} group is Electricity and Traffic.}
\label{tab:bds_app}
\begin{tabular}{llcccc}
\toprule
Comparison & Group & $n$ cells & Mean $\Delta$ & Significant cells & Mean fraction $\Delta < 0$ \\
\midrule
DLinear $-$ Ridge        & standard & 24 & $-0.56$ & 13/24 & 0.59 \\
DLinear $-$ Ridge        & large    & 16 & $-0.77$ &  9/16 & 0.61 \\
Transformer $-$ Ridge    & standard & 24 & $-2.86$ & 10/24 & 0.55 \\
Transformer $-$ Ridge    & large    & 16 & $+0.18$ &  9/16 & 0.38 \\
iTransformer $-$ Ridge   & standard & 24 & $+2.40$ & 14/24 & 0.44 \\
iTransformer $-$ Ridge   & large    & 16 & $-0.31$ &  9/16 & 0.45 \\
PatchTST $-$ Ridge       & standard & 24 & $-7.83$ & 18/24 & 0.70 \\
PatchTST $-$ Ridge       & large    &  8 & $-0.63$ &  1/8  & 0.62 \\
\bottomrule
\end{tabular}
\end{table}

On the six standard benchmarks PatchTST shows a clear negative difference (mean $\Delta = -7.83$, significant in $18$ of $24$ cells). It does leave less non-random structure in its residuals than the linear model, indicating that it captures nonlinear patterns the linear model cannot. However, on those same six datasets this additional structure does not change the accuracy ranking, on which the tuned linear model still matches or outperforms PatchTST in Table~\ref{tab:main_results}. Two further observations qualify a simple ``more capacity captures more structure'' interpretation. On Weather the difference is \emph{positive} and significant at three of four horizons, indicating that PatchTST's residuals are \emph{more} structured than the linear model's on that dataset. On the two large datasets, where PatchTST does achieve lower error, only one of eight comparisons is significant, so its advantage there is not explained by the capture of nonlinear structure.

In summary, the strongest nonlinear model does extract some patterns that the linear model does not capture, and the test detects them clearly on the six standard datasets. However, on those same datasets this additional structure does not translate into lower error. The tuned linear model still matches or outperforms the nonlinear one (Table~\ref{tab:main_results}). In other words, most of what determines accuracy on these benchmarks is the long-range linear signal documented in Appendix~\ref{sec:app_acf}, exploited through an appropriate context length as documented in Appendix~\ref{sec:app_lookback}. The additional nonlinear structure that a higher-capacity model captures is real but yields little or no accuracy benefit here. On the two large datasets, where a higher-capacity model does achieve lower error, this test indicates that its advantage arises from a source other than the capture of nonlinear structure.

\section{Limitations}
\label{sec:limitations}
Our study is limited to standard numeric long-horizon forecasting benchmarks, so the lookback-scaling, locality, and heterogeneity patterns we report should be read as findings for this regime rather than universal claims. We follow prior work in reporting point-estimate MSE; very small margins should therefore be treated as ties. Nonlinear baselines are quoted from their original publications, so comparisons reflect published configurations rather than a jointly retuned preprocessing protocol~\citep{qiu2024tfb}. Per-series search scales with channel count, which motivates our practical choice of $g_h{=}48$ and moderate $g_s$; finer-grained search may yield further gains. Finally, our claims are specific to Ridge with searched preprocessing. Applying the same preprocessing search to deeper models, and comparing all model classes under jointly tuned preprocessing, remains future work.

\section{Broader Impact}
\label{sec:broader_impact}
This work is methodological and is evaluated entirely on public numeric time-series benchmarks. It does not use human-subject data, target individual users, or study deployment in a real operational system. Its most direct positive impact is practical: closed-form Ridge models with searched preprocessing provide a strong, reproducible baseline at substantially lower compute cost than many nonlinear forecasters, making rigorous long-horizon forecasting comparisons more accessible in settings without large GPU budgets. The transparency of the resulting models also supports diagnostics: selected context length, locality, and per-series disagreement can be inspected directly and may inform the design of larger forecasters.

The main potential negative impacts are generic to forecasting research. More accurate long-horizon forecasts can be applied to sensitive operational signals, including energy, traffic, and financial time series, where misuse could affect privacy, resource allocation, infrastructure operation, or markets. We do not study such deployments, and the released pipeline is a reusable forecasting method rather than a system targeted at individuals or groups. We therefore see no additional risks specific to this work beyond these general dual-use concerns.

\section{Licenses and Terms of Use for Existing Assets}
\label{app:licenses}

We use existing public benchmark datasets, baseline numbers, and standard scientific software, and credit the original creators below. All assets are used unmodified and within their stated terms of use. Where we have verified the license against the asset's distribution channel, we name it; otherwise we point to the canonical repository, whose terms govern use.

\paragraph{Benchmark datasets.} The eight multivariate forecasting benchmarks used in Table~\ref{tab:main_results} are obtained from the public benchmark distribution accompanying~\cite{wu2021autoformer} (\url{https://github.com/thuml/Autoformer}, MIT License). Their original sources and terms are:
\begin{itemize}
    \item \textbf{ETTh1, ETTh2, ETTm1, ETTm2} (Electricity Transformer Temperature). Originally released by the authors of Informer at \url{https://github.com/zhouhaoyi/ETDataset}. Refer to the repository for the dataset's terms of use.
    \item \textbf{Electricity} (\textit{ElectricityLoadDiagrams20112014}). UCI Machine Learning Repository, \url{https://archive.ics.uci.edu/dataset/321/electricityloaddiagrams20112014}, distributed under the Creative Commons Attribution 4.0 International (CC BY 4.0) license.
    \item \textbf{Traffic} (Caltrans PeMS road-occupancy data). California Department of Transportation, \url{https://pems.dot.ca.gov/}, made publicly available by Caltrans for research and operational use.
    \item \textbf{Weather}. Max Planck Institute for Biogeochemistry, \url{https://www.bgc-jena.mpg.de/wetter/}, publicly distributed by the institute.
    \item \textbf{Exchange Rate}. Released by the authors of LSTNet at \url{https://github.com/laiguokun/multivariate-time-series-data} under the MIT License.
\end{itemize}

\paragraph{Baseline numbers and reference implementations.} The linear baseline numbers (OLS, FITS, DLinear) in Table~\ref{tab:main_results} are quoted from~\cite{toner2024analysis}. The nonlinear baseline numbers (PatchTST, iTransformer, TimeMixer, TimesNet, Autoformer) are quoted from the corresponding original publications. The reference implementations released by their authors are:
\begin{itemize}
    \item \textbf{DLinear / NLinear}~\citep{zeng2023dlinear}: \url{https://github.com/cure-lab/LTSF-Linear} (Apache License 2.0).
    \item \textbf{PatchTST}~\citep{nie2022patchtst}: \url{https://github.com/yuqinie98/PatchTST} (Apache License 2.0).
    \item \textbf{iTransformer}~\citep{liu2023itransformer}: \url{https://github.com/thuml/iTransformer} (MIT License).
    \item \textbf{TimeMixer}~\citep{wang2024timemixer}: \url{https://github.com/kwuking/TimeMixer} (Apache License 2.0).
    \item \textbf{TimesNet}~\citep{wu2022timesnet} via the Time-Series-Library: \url{https://github.com/thuml/Time-Series-Library} (MIT License).
    \item \textbf{Autoformer}~\citep{wu2021autoformer}: \url{https://github.com/thuml/Autoformer} (MIT License).
    \item \textbf{TFB}~\citep{qiu2024tfb}: \url{https://github.com/decisionintelligence/TFB} (Apache License 2.0).
\end{itemize}

\paragraph{Software libraries.} The pipeline is built on standard open-source scientific Python: PyTorch (BSD-3-Clause), NumPy (BSD-3-Clause), Pandas (BSD-3-Clause), SciPy (BSD-3-Clause), scikit-learn (BSD-3-Clause), statsmodels~\citep{seabold2010statsmodels} (BSD-3-Clause), Matplotlib (PSF-style Matplotlib license), Seaborn (BSD-3-Clause), and Optuna~\citep{akiba2019optuna} (MIT License). All libraries are used unmodified and within their respective licenses.

\end{document}